\pdfoutput=1

\documentclass[11pt]{article}

\usepackage{ACL2023}

\usepackage{times}
\usepackage{latexsym}
\usepackage{enumitem}
\usepackage[T1]{fontenc}

\usepackage[utf8]{inputenc}

\usepackage{microtype}

\usepackage{inconsolata}

\usepackage{hyperref}       
\usepackage{url}            
\usepackage{booktabs}       
\usepackage{amsfonts}       
\usepackage{nicefrac}       
\usepackage{microtype}      
\usepackage{lipsum}
\usepackage{fancyhdr}       
\usepackage{graphicx}       
\usepackage{float}
\usepackage{xcolor}
\usepackage{algorithm}
\usepackage{subfig}
\usepackage[noend]{algpseudocode}
\usepackage{stfloats}

\algnewcommand{\Inputs}[1]{%
  \State \textbf{Inputs:}
  \Statex \parbox[t]{\linewidth}{\raggedright #1}
  \vspace{0.05pt}
}
\algnewcommand{\Initialize}[1]{%
  \State \textbf{Initialize:}
  \Statex \parbox[t]{\linewidth}{\raggedright #1}
  \vspace{0.05pt}
}
\algnewcommand{\Outputs}[1]{%
  \State \textbf{Outputs:}
  \Statex \hspace*{\algorithmicindent}\parbox[t]{\linewidth}{\raggedright #1}
}
\algrenewcommand{\algorithmiccomment}[1]{\hfill\textcolor{gray}{// #1}}

\usepackage{multirow}
\usepackage{bbold}
\usepackage{amsmath}
\usepackage{url}
\usepackage{pifont} 
\usepackage{makecell}

\newcommand{\RN}[1]{%
	\textup{\lowercase\expandafter{\it \romannumeral#1}}%
}

\title{Enhancing Task Bot Engagement with Synthesized Open-Domain Dialog}

\author{\stepcounter{footnote}Miaoran Li\thanks{\ \ This work was done during an internship at Microsoft Research.} \\
  Iowa State University \\
  \texttt{limr@iastate.edu} \\\And
  Baolin Peng \\
  Microsoft Research \\
  \texttt{baolin.peng@microsoft.com} \\\And
  Michel Galley \\
  Microsoft Research \\
  \texttt{mgalley@microsoft.com} \\\AND
  Jianfeng Gao \\
  Microsoft Research \\
  \texttt{jfgao@microsoft.com} \\\And
  Zhu Zhang \\
  University of Rhode Island \\
  \texttt{zhuzhang@uri.edu}
  }

\begin{document}
\maketitle

\setlength{\abovedisplayskip}{1pt}
\setlength{\belowdisplayskip}{1pt}
\def\taskbot{TaskBot}
\def\chatbot{ChatBot}
\def\mixedbot{PivotBot}
\def\dataset{MultiWOZChat}
\def\settingone{\texttt{INITIAL}}
\def\settingtwo{\texttt{TRANSITION}}
\def\settingthree{\texttt{MULTIPLE}}

\begin{abstract}
The construction of dialog systems for various types of conversations, such as task-oriented dialog (TOD) and open-domain dialog (ODD), has been an active area of research. In order to more closely mimic human-like conversations that often involve the fusion of different dialog modes, it is important to develop systems that can effectively handle both TOD and ODD and access different knowledge sources. In this work, we present a new automatic framework to enrich TODs with synthesized ODDs. We also introduce the \mixedbot{} model, which is capable of handling both TOD and ODD modes and can access different knowledge sources to generate informative responses. Evaluation results indicate the superior ability of the proposed model to switch smoothly between TOD and ODD tasks.
\end{abstract}

\section{Introduction}
\begin{figure}[!htb]
  \centering  
  \includegraphics[width=\linewidth]{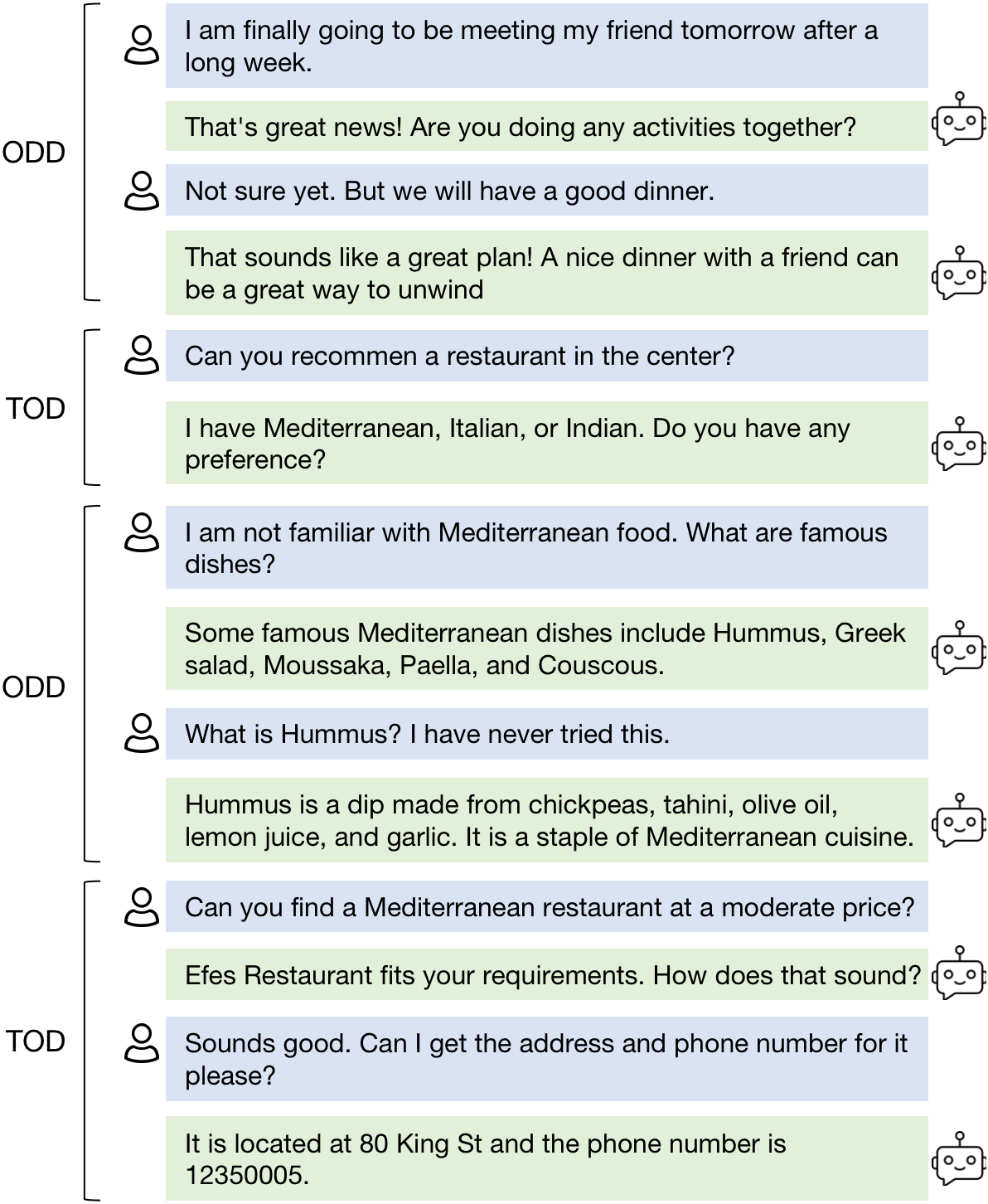}
  \caption{An example dialog that contains multiple transitions between different dialog modes.}
  \vspace{-0.5cm}
  \label{fig:intro}
\end{figure}

Task-oriented dialog (TOD) systems and open-domain dialog (ODD) systems are two active areas of Conversational AI study \cite{gao-etal-2018-neural-approaches, ni2022recent}. However, most of the existing studies model TOD and ODD systems separately, leading to a gap between the capabilities of these systems and natural human conversations. 
In real-world conversations, different dialog modes are often fused, as shown in Figure~\ref{fig:intro}. The conversation may start with casual chats and then move towards task-related requests. Along the way, the user may express interest in entities mentioned in the conversation, such as Mediterranean food in the given example, leading to a brief ODD regarding the entity of interest. The user then returns to task completion, keeping the requests in mind while maintaining a casual conversation.

To address the challenge of training dialog models to handle both TOD and ODD modes, previous research has suggested training models on mixture of TOD and ODD datasets~\cite{zhao-etal-2022-unids} or enriching existing TOD datasets by combining chitchat with TOD system responses~\cite{sun-etal-2021-adding,chen-etal-2022-ketod} or adding ODD to the beginning or end of a TOD~\cite{young2022fusing}. However, these approaches have limitations, including limited information in chitchat augmentation and a lack of explicit distinction between dialog modes. Additionally, creating new datasets through human annotation is time-consuming and expensive. While \citet{chiu-etal-2022-salesbot} have introduced a framework for automatically generating dialogs that transition from ODD to TOD, this method may not be suitable for various mode transitions and cannot simulate informative system utterances with external knowledge. 

In this work, we introduce a framework to automatically enrich TODs with synthesized ODDs. Our approach assumes that users lead conversations with explicit intentions, and that the system's objective is not only to fulfill users' requests but also to generate engaging responses on open-domain topics using external knowledge. We also consider general settings with more flexible dialog mode switches.

This paper makes the following contributions: $(\RN{1})$~We introduce a general framework for automatically enriching a TOD with knowledge-grounded ODDs and construct the \dataset{} dataset using this framework. $(\RN{2})$~We design a unified model, \mixedbot{}, that performs both TOD and ODD tasks by predicting the appropriate dialog mode and accessing knowledge sources for response generation.  $(\RN{3})$~We show experimental results that demonstrate the effectiveness of \mixedbot{} in conducting seamless conversations of both types.

\section{Proposed Framework}

\begin{figure}[!htb]
  \centering  
  \includegraphics[width=\linewidth]{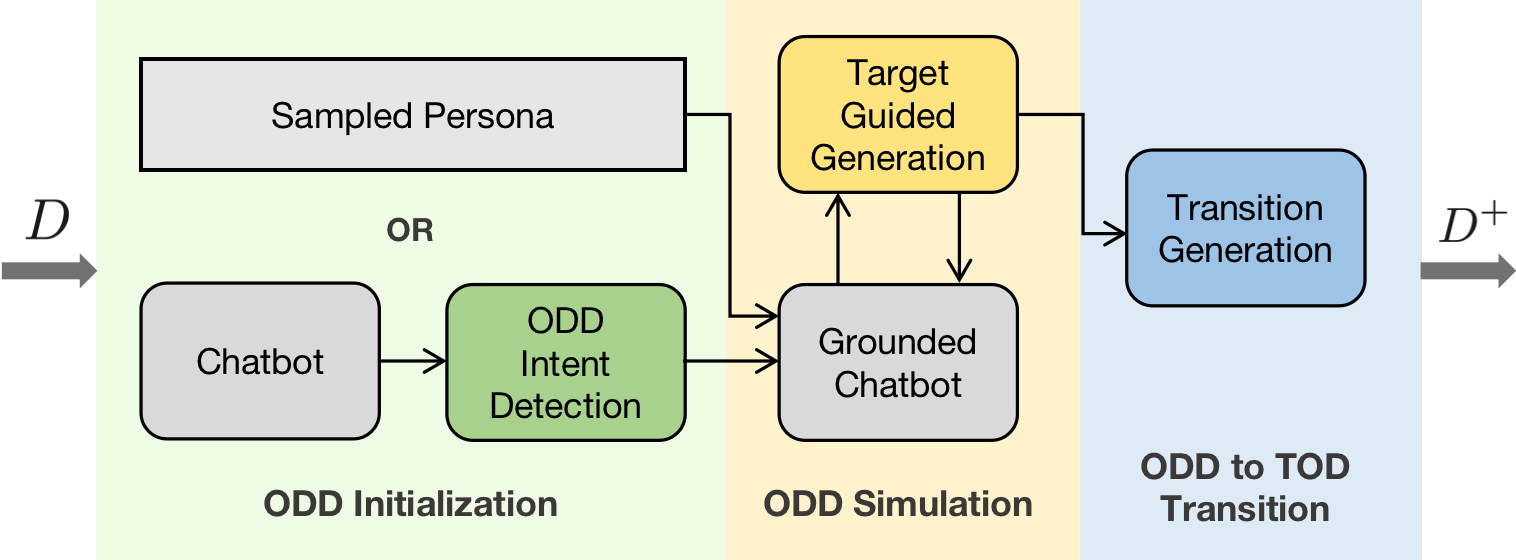}
  \caption{Framework for enriching a given TOD $D$ with ODD. The framework consists of three phases: ODD initialization, ODD simulation, and ODD-to-TOD transition. Rounded and sharp-corner boxes represent models and variables, respectively. The gray color indicates that the model is off-the-shelf. The output is the augmented dialog $D^+$.}
  \vspace{-0.2cm}
  \label{fig:framework}
\end{figure}

Figure~\ref{fig:framework} shows the proposed framework for automatically synthesizing one or more knowledge-grounded ODDs to a given TOD. The framework consists of three stages: (1) ODD initialization (2) ODD simulation, and (3) ODD to TOD transition. 
We define the following notations:
\begin{itemize}[leftmargin=0.5cm,topsep=4pt,itemsep=1pt]
    \item Denote TOD by $D\! =\! \{\boldsymbol{u}^{d_1}_1,\boldsymbol{s}^{d_1}_1,...,\boldsymbol{u}^{d_1}_{n_1},\boldsymbol{s}^{d_1}_{n_1},$ $...,\boldsymbol{u}^{d_2}_{n_1+n_2},\boldsymbol{s}^{d_2}_{n_1+n_2},...,\boldsymbol{u}^{d_N}_{n},\boldsymbol{s}^{d_N}_{n}\}$,\footnote{For settings we do not care about domains in TOD, $D$ can be simplified to $\{\boldsymbol{u}_1,\boldsymbol{s}_1,...,\boldsymbol{u}_n, \boldsymbol{s}_n\}$.} where $N$ is the number of domains in the dialog, $\boldsymbol{u}^{d_j}_i$ and $\boldsymbol{s}^{d_j}_i$ are user and system utterances at turn $i$ in domain $j$, $n_i$ is the number of turns in domain $d_{i}$, and $n$ is the total number of turns in $D$.
    \item Denote synthesized ODD by $D'\! =\! \{\boldsymbol{u}'_{1},\boldsymbol{s}'_{1},$ $...,\boldsymbol{u}'_{n'},\boldsymbol{s}'_{n}\}$, where $n'$ is the number of turns in the ODD, $\boldsymbol{u}'_t$ and $\boldsymbol{s}'_t$ represent user and system utterances at turn $t$, respectively.
\end{itemize}

\noindent
Detailed implementation of each module can be found in Appendix~\ref{implement}.

\subsection{ODD Initialization}
\label{chap:initialization}
Given a TOD $D$, we initialize the synthesized ODD $D'$ in two ways. If the ODD serves as the preface to the TOD, it is initialized by a randomly sampled user persona. If the ODD is inserted into the TOD as interludes and generated based on the TOD history, we leverage an existing chatbot to simulate a user utterance that can be inserted at a potential location. We then utilize this simulated user utterance to detect whether the user intends to have an open-domain conversation. The off-the-shelf BlenderBot model \cite{Roller2021} is used as the chatbot in the implementation. These two initialization methods are employed across diverse simulation settings (Section~\ref{settings}).

\paragraph{ODD Intent Detection}To determine the appropriate time to include an ODD during task completion, we focus on detecting the user's intent to divert the conversation from the task and discuss context-related topics.
Given a user utterance $\boldsymbol{u} = \{u_1,...,u_n\}$, where $u_i$ is the $i$-th token in the utterance, the ODD intent detection model aims to predict whether the utterance is in a TOD setting or ODD setting. The model is trained by minimizing cross-entropy loss:
\begin{align}
    \label{eq:detection}
    \mathcal{L}(\hat{I}, I) &= \sum_{i=1}^N-(\mathbb{1}(\hat{I}_i = I_i) \log(p_{\boldsymbol{\theta}}(I_i) \nonumber\\
    &+(1-\mathbb{1}(\hat{I}_i=I_i)) \log(1-p_{\boldsymbol{\theta}}(I_i)),
\end{align}
where $N$ is number of training examples, $\hat{I}_i$ and $I_i$ are predicted and ground truth intent of the $i$-th training example, $\boldsymbol{\theta}$ is the parameters of the model.


\subsection{ODD Simulation}
\label{chap:simulation}
After initializing the ODD, we use a knowledge-grounded chatbot to mimic a system with access to external knowledge and a target-guided generation model to simulate a user. In practice, we adopt the BlenderBot 2.0 model \cite{Xu2022,komeili2022internet} and BlenderBot model to simulate system and user utterances, respectively.
The ODD is considered complete if a goal $\boldsymbol{g}$ extracted from the subsequent TOD snippet is mentioned in a simulated user utterance. 

\paragraph{Target-guided Generation}To simulate the human user in the given TOD, we train a target-guided generation model that is designed to generate utterances based on the dialogue history and mention a preset target at the end of the ODD. 
The target-guided generation model is expected to generate a user utterance $\boldsymbol{u}'$ at turn $t+1$ based on a pre-determined target $\boldsymbol{g}$ and dialog context $\boldsymbol{c}$ up to turn $t$.\footnote{We conducted pilot experiments using formulations that included keyword prediction, but found not significant performance improvement. Thus, we decided to use the simplest formulation without turn-level keyword transitions.} The target is extracted from the initial user utterance of the subsequent TOD part. Given pre-determined ODD goal $\boldsymbol{g} = \{g_1,...,g_{N_g}\}$ and context $\boldsymbol{c}$, where $g_i$ is the $i$-th token in the goal, the training objective is defined as
\begin{align}
    \label{eq:utter_gen}
    \mathcal{L}_U &= \log p(\boldsymbol{u}'_{t+1}\mid\boldsymbol{g},\boldsymbol{c})\nonumber\\
    &=\sum_{i=1}^{N_u} \log p_{\boldsymbol{\theta}}(u'_{t+1,i}\mid u'_{t+1,<i}, \boldsymbol{g},\boldsymbol{c}),
\end{align}
where $\boldsymbol{\theta}$ is the set of trainable parameters in the model, $N_u$ is the target length of predicted user utterance, and $u_{t+1,<i}$ represents tokens before index $i$ of predicted user utterance at turn $t+1$.

\subsection{ODD to TOD Transition}
\label{chap:transition}
Finally, we generate a transition from the simulated ODD to the subsequent TOD to make the dialog more natural.
The goal of transition generation is to predict a system utterance that can smoothly connect the last user utterance in the ODD with the initial user utterance in the following TOD. The training objective is 
\begin{align}
    \label{eq:transition}
    \mathcal{L}_{T} &= \log p(\boldsymbol{s}'_t \mid \boldsymbol{u}'_t, \boldsymbol{u}_{t+1})\nonumber\\ 
    &= \sum_{i=1}^{N_s} \log p_{\boldsymbol{\theta}}(s'_{t,i} \mid s'_{t,<i}, \boldsymbol{u}'_t, \boldsymbol{u}_{t+1}),
\end{align}
where $\boldsymbol{u}'_t$ is the last user utterance in generated ODD, $\boldsymbol{u}_{t+1}$ is the first user utterance in the following TOD, $\boldsymbol{s}'_t$ is the transition system utterance. 

\begin{figure*}[!htb]
  \centering  
  \includegraphics[width=\textwidth]{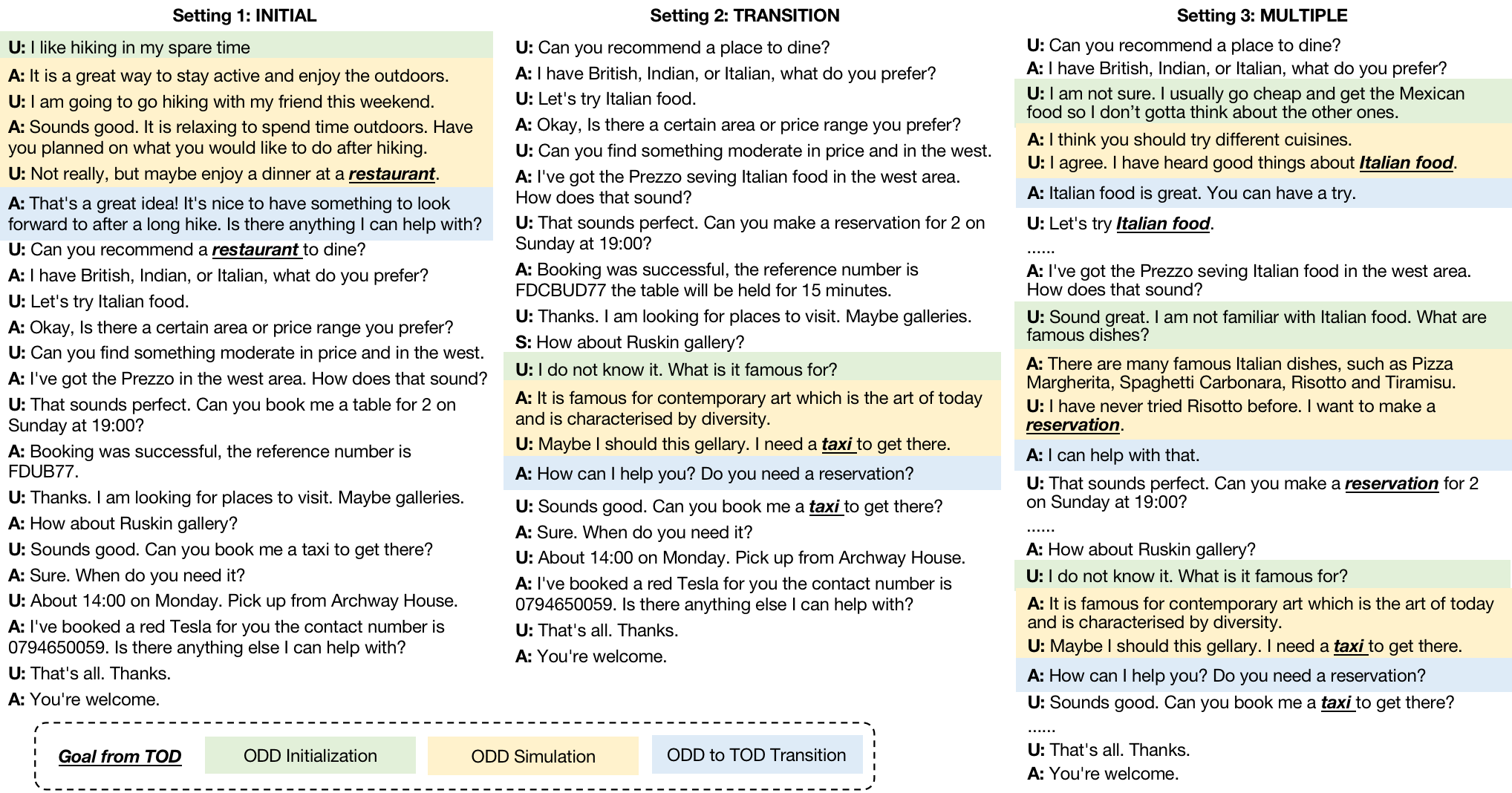}
  \vspace{-0.5cm}
  \caption{Illustration of three simulation settings. Given a TOD between a user (U) and a system agent (A), we consider three settings to synthesize ODD(s) to the TOD.
  }
  \vspace{-0.2cm}
  \label{fig:simulation}
\end{figure*}
\subsection{Simulation Settings}
\label{settings}
Inspired by previous research that aims to make dialogs more natural and engaging by adding context to a given dialog~\cite{young2022fusing} or inserting topic transition turns \cite{sevegnani-etal-2021-otters}, we consider three simulation settings: prepending an ODD to a TOD, inserting an ODD as domain transition turns, and allowing ODDs to occur at any point during task completion.
The illustration of three settings is shown in Figure~\ref{fig:simulation}.

\paragraph{Setting 1: Prepending ODD to TOD (\settingone)}
\label{setting1}
We prepend an ODD to a TOD to generate dialogs with one mode switch from ODD to TOD. We assume that users initiate the conversation by having a quick ODD and then move forward to task completion. Assuming users start with a quick ODD and then move to task completion, we initialize the ODD with a persona from a manually created persona set and use a keyword from the initial user utterance in the subsequent TOD as the goal for the synthesized ODD. Once the target is mentioned in a user utterance, the ODD simulation stops. The transition generation model is then used to connect the synthesized ODD and TOD.

\paragraph{Setting 2: Inserting ODD for Domain Transition in TOD (\settingtwo)}
\label{setting2}
To make domain transitions in TODs more natural, we insert an ODD as transition turns. Suppose a TOD $D$ contains $N$ domains, where $N\ge2$. We initialize an ODD using a chatbot after completing the conversation in domain $i$, and use intent detection model to select an utterance indicating ODD intent. The target of the ODD snippet is extracted from the first user utterance in domain $i+1$. The simulation and transition generation are similar to the previous setting. In the implementation, we only add an ODD to transition from the first domain to the second domain, and use the BlenderBot model for ODD initialization. The final dialogs contain two mode switches. 

\paragraph{Setting 3: Inserting Multiple Chitchats to Enrich TODs (\settingthree)}
\label{setting3}
In this more flexible setting, users can initiate conversations with requests and engage in small talk throughout the dialogue. The approach for generating ODDs is the same as in the \settingtwo{} setting, with the difference that we attempt to insert an ODD after each system utterance $\boldsymbol{s}_i$. This allows for multiple mode switches in the final dialogue.

\subsection{\dataset{} Dataset}
We construct \dataset{} dataset using the new framework to automatically enrich TODs from the MultiWOZ 2.1 dataset \cite{eric-etal-2020-multiwoz}. 
Table~\ref{tab:overall_stats} summarizes basic statistics of the new dataset. Focusing on the few-shot training setting, the dataset consists of 500, 198, and 1100 dialogs for the training, validation, and test sets respectively. In the \settingone{} setting, the average length of a prepended ODD is three turns, and the mean utterance length is 16.18 tokens. In the \settingtwo{} setting, the average length of a transition ODD is shorter than three turns. In the \settingthree{} setting, the average number of ODDs inserted into a TOD is four, and each ODD snippet has an average length of two turns. In the \settingtwo{} and \settingthree{} settings, the ODD durations are shorter, as they occur during task completion, and we do not want the conversation to be distracted from the task completion.
\begin{table}[!htb]
  \centering
  \resizebox{\linewidth}{!}{
  \begin{tabular}{c|c|c|c|c|c|c|c|c}
    \toprule
    Setting & Split & \makecell{Avg.\\mode\\ switch} & \makecell{Total\\ODD\\turn} & \makecell{Total\\TOD\\turn} & \makecell{Avg.\\ODD\\turn}	& \makecell{Avg.\\TOD\\turn} & \makecell{Avg.\\ODD\\length} & 
    \makecell{Avg.\\TOD\\length}\\
    \midrule
    \multirow{3}{*}{\settingone} & Train & \multirow{3}{*}{1} & 1524 & 4086 & 3.05 & 8.17 & 16.18 & 18.07 \\
    & Dev & & 565 & 1599 & 2.85 & 8.08 & 15.90 & 18.30 \\
    & Test & & 3248 & 9031 & 2.95 & 8.21 & 15.99 & 18.17\\
    \midrule
    \multirow{3}{*}{\settingtwo} & Train & \multirow{3}{*}{2} & 1301 & 4086 & 2.60 & 8.17 & 18.22 & 18.07\\
    & Dev & & 510 & 1599 & 2.58 & 8.08 & 18.26 & 18.30 \\
    & Test & & 2923 & 9031 & 2.66 & 8.21 & 18.21 & 18.17\\
    \midrule
    \multirow{3}{*}{\settingthree} & Train & 4.96 & 4356 & 4086 & 8.71 & 8.17 & 17.80 & 18.07\\
    & Dev & 4.90 & 1599 & 1599 & 8.47 & 8.08 & 17.61 & 18.30\\
    & Test & 5.11 & 9995 & 9031 & 9.87 & 8.21 & 17.82 & 18.17\\
    \bottomrule
  \end{tabular}
  }
  \vspace{-0.2cm}
  \caption{Statistics of simulated dialogs in different settings. The training, validation, and test sets comprise 500, 198, and 1100 dialogs, respectively.}
  \vspace{-0.2cm}
  \label{tab:overall_stats}
\end{table}

\section{Methodology}
\subsection{Problem Formulation} 
The full task consists of three processes: state prediction, knowledge retrieval, and knowledge-grounded response generation. We use off-the-shelf models for knowledge retrieval, which can be a database lookup or a search engine,\footnote{In the implementation, we adopted the Bing search engine.} and do not consider it as a subtask.
The full task is then divided into two subtasks: state prediction and knowledge-grounded response generation. 
In the $t$-th turn of a dialog, the model predicts the state $\boldsymbol s$ based on the dialog history $\boldsymbol h = \{\boldsymbol u_{t-k}, \boldsymbol r_{t-k},..., \boldsymbol u_t\}$, where $k$ is the size of the history window, $\boldsymbol u_i$ and $\boldsymbol r_i$ represent the user utterance and system response at the $i$-th turn, respectively. The state indicates the appropriate dialog mode and the query to obtain knowledge $\boldsymbol{k}$. The model then generates a response $\boldsymbol{r}$ based on the dialog history, predicted state and knowledge.

\subsection{\mixedbot}
\begin{figure}[!htb]
  \centering  
  \includegraphics[width=0.9\linewidth]{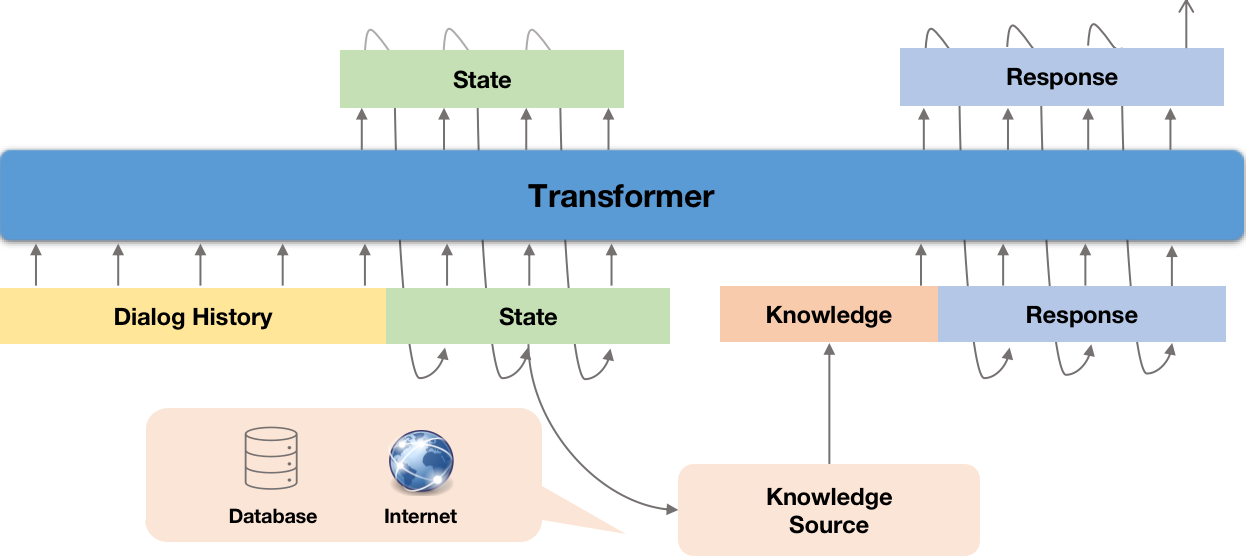}
  \caption{Overall architecture of the \mixedbot{} model}
  \vspace{-0.2cm}
  \label{fig:model}
\end{figure}
\noindent
We construct a unified model, \mixedbot{}, as shown in Figure~\ref{fig:model}. \mixedbot{} first predicts a state indicating the appropriate dialog mode and query to obtain knowledge based on the dialog history. The knowledge acquisition is completed by off-the-shelf models based on the prediction. Finally, the model performs grounded generation to generate a response.

\paragraph{State Prediction}
\begin{figure*}%
    \centering
    \subfloat[Example of performing TOD modeling]{{\includegraphics[width=0.95\columnwidth]{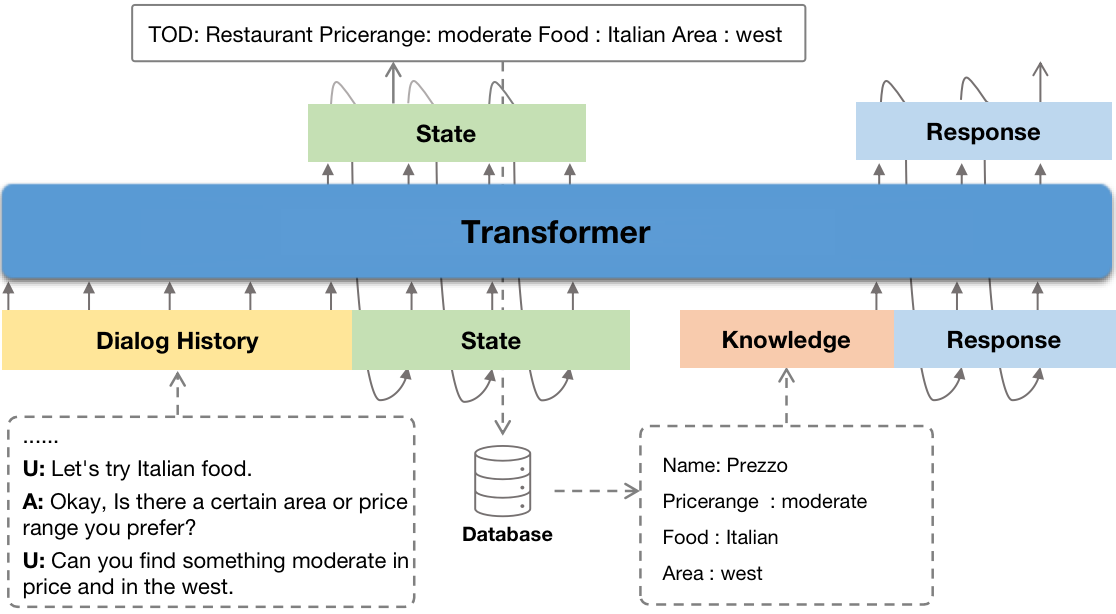} }}%
    \qquad
    \subfloat[Example of performing ODD with external knowledge]{{\includegraphics[width=0.95\columnwidth]{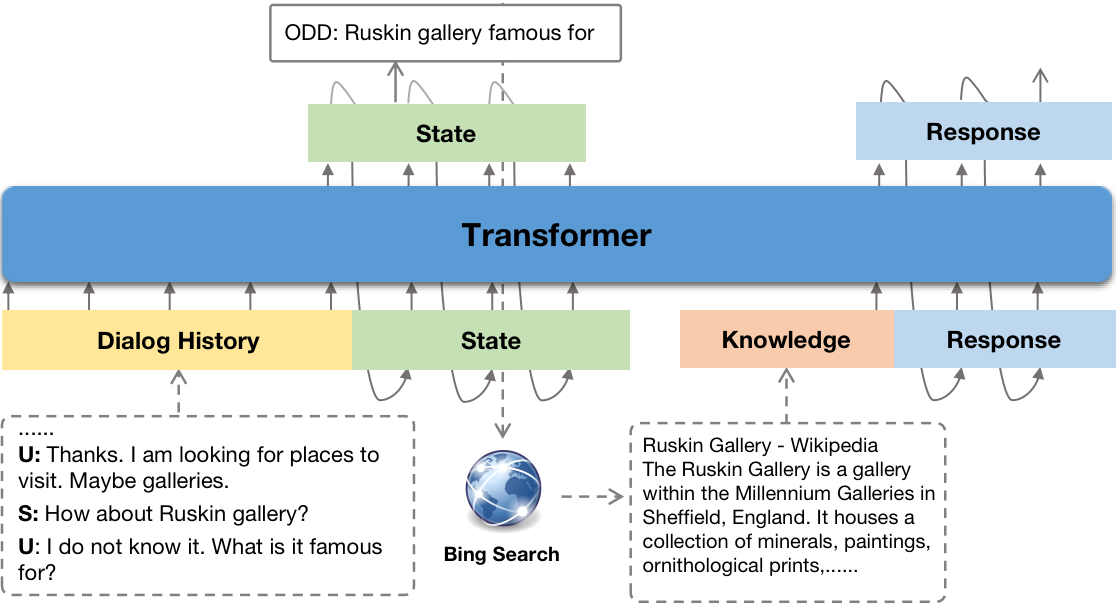} }}%
    \caption{Examples of the proposed model predicting different states}%
    \label{fig:state_example}%
\end{figure*}
State $\boldsymbol{s}$ tracks a user's goal throughout a dialog. In particular, a state $\boldsymbol{s}$ is in the form \texttt{m:q}, where \texttt{m} represents the dialog mode, and \texttt{q} stands for the query to acquire knowledge from a knowledge source. We consider two dialog modes: TOD modeling and knowledge-grounded ODD. If the model predicts performing TOD modeling, a database state is obtained from the pre-defined database using the predicted belief state (shown in Figure~\ref{fig:state_example} (a)). If the state indicates the dialog mode is ODD, external knowledge can be retrieved from the Web using the predicted search query (shown in Figure~\ref{fig:state_example} (b)). If the search query is empty, it implies that external knowledge is not needed for response generation, and the retrieved knowledge is also empty.
Given dialog history $\boldsymbol h$, the training objective of state prediction can be formulated as
\begin{equation} \label{eq:state}
    \mathcal{L}_S = \log p(\boldsymbol{s}\mid \boldsymbol{h}) = \sum_{i=1}^{N_t} \log p_{\boldsymbol{\theta}} (s_i\mid s_{<i},\boldsymbol{h}),
\end{equation}
where $\boldsymbol{\theta}a$ represents trainable parameters in the model, $N_t$ is the target length of predicted state sequence, and $s_{<i}$ denotes tokens before index $i$.

\paragraph{Grounded Generation}
System response $\boldsymbol{r} = \{r_1,r_2,...,r_{N_r}\}$ with length $N_r$ is generated grounded on dialog history $\boldsymbol{h}$, predicted state $\boldsymbol{s}$ and retrieved knowledge $\boldsymbol{k}$. In this work, the knowledge can be a database state that contains records satisfying the conditions of the belief state or retrieval results based on the search query.
The training objective is defined as
\begin{align} 
    \label{eq:response}
    \mathcal{L}_R &= \log p(\boldsymbol{r}\mid \boldsymbol{h},\boldsymbol{s}, \boldsymbol{k})\nonumber\\ 
    &= \sum_{i=1}^{N_r} \log p_{\boldsymbol{\theta}} (r_i\mid r_{<i},\boldsymbol{h}, \boldsymbol{s}, \boldsymbol{k}).
\end{align}

\paragraph{Training Objective of Full Task}
A training example consists of four components: dialog history $\boldsymbol h $, state $\boldsymbol s$, retrieved knowledge $\boldsymbol k$, and (delexicalized) dialog response $\boldsymbol r$. 
The overall training objective is 
\begin{equation}
    \mathcal{L}_{\boldsymbol{\theta}}(\mathcal{D}) = \sum_{i=1}^{N_D}(\mathcal{L}_S(\boldsymbol{x_i}) + \mathcal{L}_R(\boldsymbol{x_i})),
\end{equation}
where $\mathcal{D} = \{\boldsymbol{x}_i\}_{i=1}^{N_D}$ is the training dataset containing $N_D$ training examples. 

\section{Experiments}
\subsection{Experimental Setup}
We train models using 100, 200, and 500 dialogs and evaluate them on the entire test set. Our primary focus is evaluating the models trained in the few-shot setting, as this approach more closely reflects real-world scenarios.

\paragraph{Baselines} 
Previous studies either do not distinguish different dialog modes or only focus on social chats without external knowledge. However, our task requires models to switch between ODD and TOD modes and choose the appropriate knowledge source. To ensure a fair comparison, we train two baselines for our problem setting instead of comparing with models designed for different settings.
\begin{itemize}[leftmargin=0.5cm,topsep=4pt,itemsep=1pt]
    \item \taskbot{} serves as a baseline and is only capable of performing TOD with access to a database, which is trained solely on TOD turns in the \dataset{} dataset.
    \item \chatbot{} is a baseline model that can only perform ODD, which is trained on ODD turns in the \dataset{} dataset.
\end{itemize}
The baselines and \mixedbot{} are implemented using HuggingFace T5-base~\cite{raffel2020exploring} and GODEL~\cite{peng2022godel}. Further details of implementations can be found in Appendix~\ref{implement}.


\paragraph{Implementation}
The models are implemented using HuggingFace T5-base and GODEL. Training examples are truncated or padded to a length of 512. To ensure input strings contain dialog history and retrieved knowledge, the history is truncated on the left with a max length of 256 and consists of five utterances with a history window size of 2. AdamW optimizer~\cite{Loshchilov18} with a constant learning rate of 0.001 is used for training with a mini-batch size of 8 on a Tesla P100 for up to 15 epochs or until no validation loss decrease is observed. Each setting is evaluated eight times with random seeds.

\paragraph{Evaluation Metrics}
We evaluate the performance of the models in three settings: (1) standard TOD completion~\cite{Budzianowski18, Eric20, nekvinda-dusek-2021-shades}, (2) ODD response generation, and (3) the full task involving both TOD and ODD.

We evaluate TOD completion using four metrics:
(1)~\verb|BLEU|~\cite{Papineni02} measures the fluency of the generated responses;
(2)~\verb|Success| indicates if all requested attributes are answered;
(3)~\verb|Inform| measures whether the correct entity is provided (e.g., restaurant address);
(4)~\verb|Combine| score is an overall measure calculated as as (\verb|Inform|+\verb|Success|) $\times$ 0.5 + \verb|BLEU|.

We evaluate ODD using three metrics: 
(1)~\verb|Accuracy| measures the model's ability to predict the correct dialog mode, which can be calculated by comparing the predicted dialog mode with the ground truth mode;
(2)~\texttt{Success Rate} assesses the model's performance in state prediction at the dialog level, and measures the model's potential for success in the ODD task. It can be calculated by dividing the number of dialogs in which the model correctly predicts the dialog mode for all ODD turns by the total number of dialogs with ODD turns;
(3)~\verb|BLEU| measures the naturalness of the model's responses.

We evaluate the model's performance on the full task using \verb|BLEU|, \verb|Inform|, \verb|Success|, and \verb|Combine| score. \verb|BLEU| score is computed for all responses in the dialogs, while \texttt{Inform} and \texttt{Success} metrics are limited to dialogs that succeed in both TOD modeling and ODD tasks. The potential success of the ODD task is used as an indicator, and \texttt{Inform} and \texttt{Success} are computed for dialogs where the dialog mode predictions for all ODD turns are accurate.

\paragraph{Human Evaluation Setup} 
We conducted two-phase human evaluation. In the first stage, we hired Amazon Mechanical Turk workers to interact with three models: \taskbot{} with T5 as the backbone (T5-\taskbot), \mixedbot{} with T5 as the backbone (T5-\mixedbot), and \mixedbot{} with GODEL as the backbone (GODEL-\mixedbot). The workers were provided with information-seeking goals from the MultiWOZ 2.1 dataset and allowed to chat freely with the models to complete the goals. After each conversation, workers rated the  appropriateness~\cite{moghe-etal-2018-towards} and engagingness~\cite{zhang-etal-2018-personalizing} of the model's responses on a 5-point Likert scale and indicated if all requests were completed. Appropriateness assesses the model's ability to understand users' utterances and requests and provide reasonable responses, while engagingness evaluates whether the model generates engaging responses and facilitates smooth conversation flow for users.

To ensure the quality of interactions during the first stage, we employed onboarding tasks with simplified information-seeking goals. Only qualified workers who can complete the onboarding task were granted access to the main task with higher rewards. Both the onboarding and main task submissions were required to cover all necessary keywords and phrases, and each utterance had to be meaningful and not excessively brief. Additionally, we implemented manual checks on randomly sampled submissions to maintain the quality of collected results.

In the second stage, we conducted a static evaluation of the dialogs collected in the previous phase. Each worker was presented with a pair of dialogs, one produced by T5-\taskbot{} and the other by T5-\mixedbot{}, or one produced by T5-\mixedbot{} and the other by GODEL-\mixedbot{}, and was asked to choose the better dialog based on the system performance. Then workers rated the appropriateness and engagingness of each system's utterances in the dialogs using a 5-point Likert scale.

\subsection{Automatic Evaluation Results}
We present the results for models trained in the few-shot setting using 100 training dialogs with the GODEL backbone.\footnote{We also evaluated the models using the T5-base backbone and found that models with the GODEL backbone outperform those based on T5-base, with statistically significant performance differences.}
For the full task evaluation, we only report the combined score. The evaluation results using 200 and 500 training dialogs are in Appendix~\ref{full_eval}.

\paragraph{\settingone{} Setting Evaluation}
\begin{table*}[!htb]
  \centering
  \resizebox{\textwidth}{!}{
  \begin{tabular}{c|c|cccc|ccc}
    \toprule
    \multirow{2}{*}{Model} & Full Task & \multicolumn{4}{c|}{TOD Evaluation} & \multicolumn{3}{c}{ODD Evaluation}\\
    & Combined & BLEU & Success & Inform & Combined & Accuracy & Success Rate & BLEU\\
    \midrule
    \taskbot & 12.26(0.43) & \textbf{15.00}(0.57) & 37.43(4.02) & 52.61(4.26) & 60.01(4.41) & 0.00(0.00) & 0.00(0.00) & 1.33(0.22)\\
    \chatbot & 7.98(0.25) & 0.97(0.16) & 0.60(0.00) & 10.70(0.00) & 6.62(0.16) & \textbf{99.93}(0.05) & \textbf{99.79}(0.16) & 6.43(0.51)\\
    \mixedbot & \textbf{58.06}(5.15) & 14.90(0.58) & \textbf{38.66}(5.22) & \textbf{53.55}(5.62) & \textbf{61.01}(5.48) & 98.90(0.45) & 97.35(0.98) & \textbf{6.82}(0.41)\\
    \bottomrule
  \end{tabular}}
  \vspace{-0.2cm}
    \caption{End-to-end evaluation in the \settingone{} setting. Mean values and standard deviations (in parentheses) are reported.}
    \vspace{-0.2cm}
    \label{tab:setting1}
\end{table*}
Table~\ref{tab:setting1} shows the evaluation results in the \settingone{} setting. 
\mixedbot{} significantly outperforms the baseline models in the full task evaluation, demonstrating the importance of incorporating different dialog modes.
\mixedbot{} also slightly outperforms \taskbot{} in the TOD task in terms of the \verb|Combined| score. This suggests that the ability to handle both TOD and ODD tasks with appropriate dialog modes and knowledge sources is critical for \mixedbot{} to excel in the full task.
While \chatbot{} cannot provide requested attributes or entities, it performs better than other models in predicting the dialog mode in the ODD evaluation setting. Though \mixedbot{} cannot beat \chatbot{} in the ODD evaluation, it achieves comparable results while generating more fluent responses and simultaneously handling task completion. 

\paragraph{\settingtwo{} Setting Evaluation}
\begin{table*}[!htb]
  \centering
  \resizebox{\textwidth}{!}{
  \begin{tabular}{c|c|cccc|ccc}
    \toprule
    \multirow{2}{*}{Model} & Full Task & \multicolumn{4}{c|}{TOD Evaluation} & \multicolumn{3}{c}{ODD Evaluation}\\
    & Combined & BLEU & Success & Inform & Combined & Accuracy & Success Rate & BLEU\\
    \midrule
    \taskbot & 12.37(0.36) & \textbf{15.02}(0.46) & \textbf{35.43}(4.14) & \textbf{50.56}(5.35) & \textbf{58.02}(5.03) & 0.00(0.00) & 0.00(0.00) & 1.22(0.12)\\
    \chatbot & 7.88(0.13) & 1.22(0.16) & 0.60(0.00) & 10.70(0.00) & 6.87(0.16) & \textbf{100.00}(0.01) & \textbf{99.99}(0.03) & \textbf{5.35}(0.18)\\
    \mixedbot & \textbf{49.58}(7.13) & 14.92(0.64) & 33.49(6.13) & 47.06(8.21) & 55.19(7.56) & 96.17(0.64) & 90.00(1.72) & 4.97(0.28)\\
    \bottomrule
  \end{tabular}}
  \vspace{-0.2cm}
  \caption{End-to-end evaluation in the \settingtwo{} setting. Mean values and standard deviations (in parentheses) are reported.}
  \vspace{-0.2cm}
  \label{tab:setting2}
\end{table*}
Table~\ref{tab:setting2} contains evaluation results in the \settingtwo{} setting. \mixedbot{} performs significantly better than baselines in the full task. \taskbot{} slightly outperforms \mixedbot{} in the TOD modeling task. \chatbot{} still achieves the best performance in the ODD task. Though \mixedbot{} cannot perform better than baselines in single task evaluation, it can obtain comparable results with the specialist baselines. The gap between \chatbot{} and \mixedbot{} in success rate is more obvious, indicating that it is more challenging for the model to learn both dialog modes simultaneously and accurately predict the mode when the mode switches in dialogs become more complex. 

\paragraph{\settingthree{} Setting Evaluation}

\begin{table*}[!htb]
  \centering
  \resizebox{\textwidth}{!}{
  \begin{tabular}{c|c|cccc|ccc}
    \toprule
    \multirow{2}{*}{Model} & Full Task & \multicolumn{4}{c|}{TOD Evaluation} & \multicolumn{3}{c}{ODD Evaluation}\\
    & Combined & BLEU & Success & Inform & Combined & Accuracy & Success Rate & BLEU\\
    \midrule
    \taskbot & 8.10(0.27) & \textbf{14.79}(0.48) & 34.74(5.29) & \textbf{50.16}(6.69) & 57.24(6.11) & 0.00(0.00) & 0.00(0.00) & 0.93(0.07)\\
    \chatbot & 8.76(0.29) & 1.14(0.09) & 0.60(0.00) & 10.70(0.00) & 6.79(0.09) & \textbf{100.00}(0.00) & \textbf{100.00}(0.00) & \textbf{5.05}(0.48)\\
    \mixedbot & \textbf{42.43}(3.23) & 14.77(0.65) & \textbf{35.75}(3.13) & 49.76(4.32) & \textbf{57.52}(3.89) & 96.66(0.28) & 74.39(2.15) & 4.97(0.42)\\    
    \bottomrule
  \end{tabular}}
  \vspace{-0.2cm}
  \caption{End-to-end evaluation in the \settingthree{} setting. Mean values and standard deviations (in parentheses) are reported.}
  \vspace{-0.2cm}
  \label{tab:setting3}
\end{table*}
The evaluation results in the \settingthree{} setting are presented in Table~\ref{tab:setting3}. In the full task evaluation, \mixedbot{} remains the best-performing model. The performance of \taskbot{} and \mixedbot{} is comparable in the TOD task.
However, in the ODD task evaluation, while \mixedbot{}'s turn-level prediction accuracy does not significantly decrease, the model is more likely to fail in the ODD task at the dialog level due to the increased number of ODD turns and more complex mode switches within a dialog.

\paragraph{Cross-Setting Evaluation}
\vspace{-0.2cm}
\begin{table}[!htb]
    \centering
    \resizebox{\columnwidth}{!}{
    \begin{tabular}{c|c|c|c}
        \toprule
        \multirow{2}{*}{\makecell{Training\\Setting}} & \multicolumn{3}{c}{Evaluation Setting}\\
        \cline{2-4}
        & \settingone & \settingtwo & \settingthree\\
        \midrule
        \settingone & \textbf{58.06}(5.15) & 12.12(0.42) & 8.54(0.38)\\
        \settingtwo & 22.80(10.99) & 49.58(7.13) & 22.26(2.23)\\
        \settingthree & 49.69(6.20) & \textbf{51.91}(4.28) & \textbf{42.43}(3.23)\\
        \bottomrule
    \end{tabular}}
    \vspace{-0.2cm}
    \caption{End-to-end cross setting evaluation results. Mean values and standard deviations (in parentheses) of the \texttt{Combined} score for \mixedbot{} models trained in different settings are reported.}
    \vspace{-0.2cm}
    \label{tab:overall_cross}
\end{table}
Table~\ref{tab:overall_cross} contains the \verb|Combined| scores of \mixedbot{} trained in each setting evaluated in all three settings, allowing us to examine the relationships among the different settings. 
The model trained in the \settingone{} setting performs best in that same evaluation setting. The model trained in the \settingtwo{} setting obtains comparable performance with the model in the \settingthree{} setting in the \settingtwo{} evaluation setting but struggles in the other two evaluation settings. The model trained in the \settingthree{} setting obtains the highest \verb|Combined| scores in the other two evaluation settings, indicating its ability to generalize well to different settings.

\subsection{Human Evaluation Results}
\begin{table}[!htb]
    \centering
    \resizebox{\columnwidth}{!}{
    \begin{tabular}{c|c|c|c}
        \toprule
        \multirow{2}{*}{}& T5-\taskbot & T5-\mixedbot & GODEL-\mixedbot\\
        \midrule
        Success & $0.99(0.10)$ \hfill & $1.00(0.07)$ \hfill & $\textbf{1.00}(0.00)$ \hfill\\
        Appropriateness & $4.10(1.11)$ \hfill & $4.27(1.00)$ \hfill & $\textbf{4.35}(0.01)$ \hfill\\
        Engagingness & $4.09(1.13)$ \hfill & $4.31(0.88)$ \hfill & $\textbf{4.44}(0.71)$ \hfill\\
        \bottomrule
    \end{tabular}}
    \vspace{-0.2cm}
    \caption{Results of the first phrase of human evaluation. Mean values and standard deviations (in parentheses) are reported. \texttt{Success} is measured in binary scale, while \texttt{Appropriate} and \texttt{Engagingness} are measured on a 5-point Likert scale.}
    \vspace{-0.2cm}
    \label{tab:human-eval1}
\end{table}
\begin{table}[!htb]
    \centering
    \resizebox{0.95\columnwidth}{!}{
    \begin{tabular}{c|p{1.4cm}p{1.4cm}p{1.4cm}}
        \toprule
        \multirow{2}{*}{}& \multicolumn{3}{c}{T5-\mixedbot{} vs. T5-\taskbot}\\
        \cline{2-4}
        & \hspace{0.18cm} Win & \hspace{0.18cm} Tie & \hspace{0.18cm} Loss\\
        \midrule
        Overall & \hspace{0.1cm} $51.52^{*}$ \hfill & \hspace{0.1cm} $17.68$ \hfill & \hspace{0.1cm} $30.81^{*}$ \hfill\\
        Appropriateness & \hspace{0.1cm} $50.51^{**}$ \hfill & \hspace{0.1cm} $36.87$ \hfill & \hspace{0.1cm} $12.63^{**}$ \hfill\\
        Engagingness & \hspace{0.1cm} $50.51^{**}$ \hfill & \hspace{0.1cm} $30.30$ \hfill & \hspace{0.1cm} $19.19^{**}$ \hfill\\
        \midrule
        \midrule
        \multirow{2}{*}{}& \multicolumn{3}{c}{GODEL-\mixedbot{} vs. T5-\mixedbot}\\
        \cline{2-4}
        & \hspace{0.18cm} Win & \hspace{0.18cm} Tie & \hspace{0.18cm} Loss\\
        \midrule
        Overall & \hspace{0.1cm} $44.72$ \hfill & \hspace{0.1cm} $23.62$ \hfill & \hspace{0.1cm} $31.66$ \hfill\\
        Appropriateness & \hspace{0.1cm} $43.94^{**}$ \hfill & \hspace{0.1cm} $43.22$ \hfill & \hspace{0.1cm} $13.07^{**}$ \hfill\\
        Engagingness & \hspace{0.1cm} $53.77^{**}$ \hfill & \hspace{0.1cm} $34.17$ \hfill & \hspace{0.1cm} $12.06^{**}$ \hfill\\
        \bottomrule
    \end{tabular}}
    \vspace{-0.2cm}
    \caption{Results of the second phrase of human evaluation. "Overall" stands for the dialog-level evaluation results. "Win" (or "Loss") refers to the percentage of cases where T5-\mixedbot{} (in the upper section) and GODEL-\mixedbot{} (in the lower section) wins (or loses). $^*$ denotes p-values of less than 0.05 and $^{**}$ represents p-values of less than 0.01.}
    \label{tab:human-eval2}
\end{table}
In the first phase, we collected 200 dialogs for each model. To make the evaluation task more manageable for the workers, we only sampled information-seeking goals involving a single domain, which may have made it easier for the models to fulfill all users' requests. The results are shown in Table~\ref{tab:human-eval1}. 
Consistent with the automatic evaluation, both \taskbot{} and \mixedbot{} can complete users' requests, with \mixedbot{} excelling in generating engaging and suitable responses. The GODEL backbone further enhances \mixedbot's engagingness.

In the second phase, we conducted pairwise comparisons of the models' performance and present the results in Table~\ref{tab:human-eval2}. 
Notably, there are fewer ties in overall performance comparisons than in evaluations of appropriateness and engagingness.
This could be because pairwise comparisons provide evaluators with a clearer choice, while evaluating appropriateness and engagingness could be more subjective. Factors like dialogue length and quality may influence evaluators' overall performance judgments, whereas appropriateness and engagingness are likely assessed solely on the model's merits.

\subsection{Case Study}
In Table~\ref{tab:casestudy}, we present example user utterances and the corresponding responses generated by different models. During the TOD turns, \chatbot{} exhibits limitations in providing valuable information on trains to the user, while \taskbot and \mixedbot{} can ask follow-up questions to effectively refine the search and provide information on satisfied entities. In the ODD example, \taskbot{} falls short in engaging in social chats with the user, restricting its interactions solely to assisting in the task of train ticket booking. In contrast, both \chatbot{} and \mixedbot{} respond informatively and engagingly in such scenarios.
\begin{table*}[!htp]
  \resizebox{\textwidth}{!}{
  \begin{tabular}{c|l|l}
    \toprule
    \multirow{5}{*}{TOD} & \multicolumn{2}{l}{\textbf{U}: I need to book a train leaving Cambridge on Thursday can you help me?}\\
    \cline{2-3}
    & \chatbot & I can help you book a train leaving Cambridge on Thursday.\\
    & \taskbot & Sure. Where will you be heading to?\\
    & \mixedbot & Where would you like to arrive at?\\
    \cline{2-3}
    &Reference& Absolutely! To help narrow your search, where would you like to arrive at?\\
    \midrule
    \multirow{5}{*}{TOD} & \multicolumn{2}{l}{\textbf{U}: Can you find me one that will arrive in Norwich please?}\\
    \cline{2-3}
    & \chatbot & I can find one that will arrive in Norwich for you.\\
    & \taskbot & [train\_id] arrives at [value\_time]. would you like me to book that for you?\\
    & \mixedbot & There is [train\_id] arriving at [value\_time]. would you like me to book that for you?\\
    \cline{2-3}
    &Reference& I have a train that arrives at [value\_time], if that's acceptable?\\
    \midrule
    \multirow{5}{*}{ODD} & \multicolumn{2}{l}{\textbf{U}: I have been to Norwich a few times. It is beautiful. I hope to go again.}\\
    \cline{2-3}
    & \chatbot & I've never been there, but I've always wanted to go someday. Have you ever been to other place in Norfolk, England?\\
    & \taskbot & What time do you want to go? [train\_id] arrives at [value\_time]. Is it suitable?\\
    & \mixedbot & I've never been there. I know it is a cathedral city in Norfolk, England.\\
    \cline{2-3}
    &Reference& I've never been there, but it's on my list of places to visit.\\
    \bottomrule
  \end{tabular}}
  \caption{Comparison of (delexicalized) responses generated by different models. "Reference" denotes the responses in \dataset{}.}
  \label{tab:casestudy}
\end{table*}

\section{Related Work}
\paragraph{Dialog Systems for Fused Task of ODD and TOD}
Several previous works have addressed the challenge of constructing dialog systems that can handle multiple dialog modes.
Some work focused on constructing systems that independently model different dialog skills or training dialog models on mixture of TOD and ODD datasets to enable it to switch between conversation styles~\cite{madotto2020attention,lin2021adapter} .  
Other approaches have involved constructing new datasets for mixed settings, by adding ODD utterances to system utterances in TODs \cite{zhao2017generative, sun-etal-2021-adding, chen-etal-2022-ketod}, or enriching TODs with human-annotated ODD snippets to include one mode transition within an augmented dialog \cite{young2022fusing}.
To reduce the need for human involvement in dataset construction, \citet{chiu-etal-2022-salesbot} proposed a framework for automatically generating dialogs that transition from ODD to TOD with a simulated user and simulated salesperson, assuming that users do not explicitly state their intentions and that the system must detect and respond to these intentions. 

\paragraph{Target-guided Generation for ODDs}
Some previous work \cite{xing2017topic, lian2019learning, Ling2021ContextControlledTN} focused on guiding the conversation generation in a short-term, while others studied the multi-turn target-guided process of conversations. \citet{Tang2019Target} proposed the task of target-guided open-domain conversation where the model leads the conversation from a random initial topic to a target word. \citet{qin2020dynamic} improved the previous work by constraining candidate keywords and augmenting responses with predicted keywords. \citet{kishinami2022target} modified the previous task setting and focused on evaluating the ability of a model to plan a target-oriented conversation. Researchers also considered actively leading a conversation to a target by incorporating knowledge graphs \cite{wu-etal-2019-proactive, xu2020knowledge, zhong2021keyword}.

\section{Conclusion and Future Work}
This paper introduces an easily-implemented and generalizable framework for enriching a TOD with ODDs in different settings. A unified model, \mixedbot{}, with both TOD and ODD dialog modes is designed. Evaluation results demonstrate the effectiveness of the proposed model and the significance of integrating multiple dialog modes for generating appropriate and engaging responses.

Future work on the data simulation can involve integrating external knowledge, such as knowledge graphs and personality traits, and exploring alternative guided generation methods to improve the consistency and control of the generated ODDs. 
To optimize the knowledge retrieval process, training a more efficient retrieval and selection model can be considered. Additionally, creating a system with comprehensive capabilities, including recommendation and personalization, would enhance its suitability for real-world applications.

\section{Ethical Considerations}
The \dataset{} dataset was created using BlenderBot models with safety controls to simulate ODDs and MultiWOZ 2.1 for TODs to exclude harmful dialogs. However, existing chatbots may still employ unsafe language, and pre-trained language models may have encountered text with social bias or toxicity, potentially leading to offensive responses from the \mixedbot{} model. Additionally, off-the-shelf chatbots might generate hallucinatory content, reducing the reliability of \mixedbot{}'s responses. Future work should prioritize exploring better safety measures and enhancing response accuracy.


\bibliography{custom}
\bibliographystyle{acl_natbib}

\clearpage
\appendix

\appendix
\section{Proposed framework}
\label{implement}
\vspace{-0.1cm}
\subsection{ODD Intent Detection} 
The detection model is implemented using HuggingFace BERT-base~\cite{devlin-etal-2019-bert} model and is trained on a combination of four datasets: MultiWOZ 2.1, ConvAI2\cite{Dinan2019TheSC}, FusedChat (with pretended ODDs), and Wizard of Wikipedia (WoW)~\cite{Dinan2019Wizard}, with equal numbers of TOD and ODD turns for balance.
\vspace{-0.6cm}
\subsection{Target-guided Generation}
\paragraph{MultiWOZ target candidate} 
\label{target_construction}
We consider values of 8 slots in the MultiWOZ 2.1 dataset as potential targets. These slots are \verb|name|, \verb|area|, \verb|pricerange|, \verb|type|, \verb|departure|, \verb|destination|, \verb|department|, and \verb|day|. The values can be represented as nouns, adjectives, or phrases.
\vspace{-0.1cm}
\paragraph{Training}
\label{sec:train_constrained}
We train the distilled BlenderBot on three datasets (FusedChat, WoW, ConvAI2) to generate diverse user utterances.
We use a keyword extraction method~\cite{Tang2019Target} to set target for ODDs in WoW and ConvAI2, and extract a target from the initial user utterance of the TOD part for the prepended ODDs from FusedChat.
\vspace{-0.1cm}
\paragraph{Inference}
We use the trained target-guided generation model to simulate the user in ODD and extract the goal $\boldsymbol{g}$ from the given TOD using the set of candidate targets from MultiWOZ 2.1.
\vspace{-0.1cm}

\subsection{Transition Generation} 
The implementation is based on the HuggingFace T5-base~\cite{raffel2020exploring} model. The training datasets are the same as Sec.\ref{sec:train_constrained}. A training example consists of user utterances at turn $t$ and $t+1$ and system response at turn $t$. 
\vspace{-0.1cm}
\section{Automatic Evaluation Results}
\label{full_eval}

\paragraph{\settingone{} Setting Evaluation} 
Table~\ref{tab:overall_1} and~\ref{tab:TOD+chat_1} show evaluation results in the \settingtwo{} setting. 
As the number of training dialogs increases, all models show improvement. \chatbot{} and \mixedbot{} models improve in generating fluent ODD responses, while \taskbot{} focuses more on TOD modeling and fails to respond appropriately to ODDs.

\begin{table}
    \centering
    \resizebox{\columnwidth}{!}{
    \begin{tabular}{c|c|cccc}
        \toprule
        \multirow{2}{*}{\makecell{\# Training\\dialogs}} & \multirow{2}{*}{Model} & \multicolumn{4}{c}{Full Task Evaluation} \\
        & & BLEU & Success & Inform & Combined\\
        \midrule
        \multirow{3}{*}{200} & \taskbot & $13.34 (0.22)^{**}$ & $0.00 (0.00)$ & $0.00 (0.00)$ & $13.34 (0.22)^{**}$\\
        & \chatbot & $2.56 (0.14)^{**}$ & $0.60 (0.00)$ & $10.71 (0.03)$ & $8.22 (0.15)^{**}$\\
        & \mixedbot & $\textbf{14.53} (0.18)^{**}$ & $\textbf{40.66} (1.81)^{**}$ & $\textbf{52.74} (2.70)^{**}$ & $\textbf{61.23} (2.24)^{**}$\\
        \midrule
        \multirow{3}{*}{500} & \taskbot & $14.41 (0.25)^{**}$ & $0.00 (0.00)$ & $0.00 (0.00)$ & $14.41 (0.25)^{**}$\\
        & \chatbot & $2.92 (0.09)^{**}$ & $0.60 (0.00)$ & $10.70 (0.00)$ & $8.57 (0.09)^{**}$\\
        & \mixedbot & $\textbf{15.76} (0.20)^{**}$ & $\textbf{42.45} (2.33)^*$ & $\textbf{53.79} (3.26)^*$ & $\textbf{63.88} (2.62)^*$\\
        \bottomrule
    \end{tabular}}
    \vspace{-0.1cm}
    \caption{End-to-end full task evaluation using GODEL as backbone in \settingone{} setting. Statistically significant differences exist between GODEL-based and T5-based models (*p<0.05, **p<0.01).}
    \vspace{-0.1cm}
    \label{tab:overall_1}
\end{table}
\begin{table}
    \centering
    \resizebox{\columnwidth}{!}{
    \begin{tabular}{c|c|cccc}
        \toprule
        \multirow{2}{*}{\makecell{\# Training\\dialogs}} & \multirow{2}{*}{Model} & \multicolumn{4}{c}{Full Task Evaluation} \\
        & & BLEU & Success & Inform & Combined\\
        \midrule
        \multirow{3}{*}{200} & \taskbot & $13.49 (0.15)^{**}$ & $0.00 (0.00)$ & $0.00 (0.00)$ & $13.49 (0.15)^{**}$\\
        & \chatbot & $2.42 (0.14)^{**}$ & $0.60 (0.00)$ & $10.70 (0.00)$ & $8.08 (0.14)^{**}$\\
        & \mixedbot & $\textbf{14.27} (0.31)^{**}$ & $\textbf{32.75} (5.67)$ & $\textbf{42.54} (7.26)$ & $\textbf{51.92} (6.53)$\\
        \midrule
        \multirow{3}{*}{500} & \taskbot & $14.49 (0.26)^{**}$ & $0.00 (0.00)$ & $0.00 (0.00)$ & $14.49 (0.26)^{**}$\\
        & \chatbot & $2.63 (0.06)^{**}$ & $0.60 (0.00)$ & $10.70 (0.00)$ & $8.28 (0.06)^{**}$\\
        & \mixedbot & $\textbf{15.49} (0.37)^{**}$ & $\textbf{41.39} (1.73)^{**}$ & $\textbf{51.65}  (2.30)^*$ & $\textbf{62.01} (2.11)^{**}$\\
        \bottomrule
    \end{tabular}}
    \vspace{-0.2cm}
    \caption{End-to-end full task evaluation using GODEL as backbone in \settingtwo{} setting. 
    Statistically significant differences exist between GODEL-based and T5-based models (*p<0.05, **p<0.01).
    }
    \vspace{-0.2cm}
    \label{tab:overall_2}
\end{table}


\paragraph{\settingtwo{} Setting Evaluation}
Table~\ref{tab:overall_2} and~\ref{tab:TOD+chat_2} contain evaluation results in the \settingtwo{} setting. 
Performance improvements can be observed for all models with an increase in training dialogs.
In addition, the response quality improves for both \chatbot{} and \mixedbot{}, and \mixedbot{} shows better ability to choose appropriate dialog modes.

\paragraph{\settingthree{} Setting Evaluation}
The evaluation results in the \settingthree{} setting, shown in Table~\ref{tab:overall_3} and ~\ref{tab:TOD+chat_3}, are consistent with the results in the previous settings. 
The \mixedbot{} model improves its ability to make more accurate predictions with an increase in the number of training dialogs.
\begin{table}
    \centering
    \resizebox{\columnwidth}{!}{
    \begin{tabular}{c|c|cccc}
        \toprule
        \multirow{2}{*}{\makecell{\# Training\\dialogs}} & \multirow{2}{*}{Model} & \multicolumn{4}{c}{Full Task Evaluation} \\
        & & BLEU & Success & Inform & Combined\\
        \midrule
        \multirow{3}{*}{200} & \taskbot & $8.90 (0.35)^{**}$ & $0.00 (0.00)$ & $0.00 (0.00)$ & $8.90 (0.35)^{**}$\\
        & \chatbot & $3.72 (0.22)^{**}$ & $0.60 (0.00)$ & $10.70 (0.00)$ & $9.37 (0.22)^{**}$\\
        & \mixedbot & $\textbf{11.43} (0.18)^{**}$ & $\textbf{29.10} (4.51)$ & $\textbf{38.54} (4.83)$ & $\textbf{45.25} (4.64)$\\
        \midrule
        \multirow{3}{*}{500} & \taskbot & $9.8 (0.18)^{**}$ & $0.00 (0.00)$ & $0.00 (0.00)$ & $9.80 (0.18)^{**}$\\
        & \chatbot & $4.19 (0.08)^{**}$ & $0.60 (0.00)$ & $10.70 (0.00)$ & $9.84 (0.08)^{**}$\\
        & \mixedbot & $\textbf{12.66} (0.12)^{**}$ & $\textbf{37.54} (4.09)^{**}$ & $\textbf{47.96} (5.43)^*$ & $\textbf{55.40} (4.65)^{**}$\\
        \bottomrule
    \end{tabular}}
    \vspace{-0.2cm}
    \caption{End-to-end full task evaluation using GODEL as backbone in \settingthree{} setting. 
    Statistically significant differences exist between GODEL-based and T5-based models. (*p<0.05, **p<0.01).
    }
    \vspace{-0.2cm}
    \label{tab:overall_3}
\end{table}

\paragraph{Cross-Setting Evaluation}
Table~\ref{tab:cross-full} and Table~\ref{tab:cross-tod-odd} present the cross-setting evaluation results. 
With more training dialogs, models show performance improvement in all evaluation settings. The model trained in the \settingthree{} setting demonstrates the ability to generalize well and obtains the highest (or comparable) scores in all settings.
\begin{table}[!b]
    \centering
    \resizebox{\columnwidth}{!}{
    \begin{tabular}{c|c|c|cccc}
        \toprule
        \multirow{2}{*}{\makecell{Evaluation\\setting}} & \multirow{2}{*}{\makecell{Training\\setting}} & \multirow{2}{*}{\makecell{\# Training\\dialogs}} & \multicolumn{4}{c}{Full Task Evaluation} \\
        & & & BLEU & Success & Inform & Combined\\
        \midrule
        \multirow{3}{*}{\settingone} 
        & \settingone & \multirow{3}{*}{500} & \textbf{15.76}(0.20) & \textbf{42.45}(2.33) & \textbf{53.79}(3.26) & \textbf{63.88}(2.62)\\
        & \settingtwo & & 15.24 (0.24) & 31.65 (8.13) & 39.93 (10.41) & 51.03 (9.41)\\
        & \settingthree & & 15.15 (0.20) & 35.84 (4.08) & 45.73 (5.64) & 55.93 (4.78)\\
        \midrule
        \multirow{3}{*}{\settingtwo} 
        & \settingone & \multirow{3}{*}{500} & 14.17 (0.33) & 1.52 (1.24) & 2.11 (1.72) & 15.99 (1.62)\\
        & \settingtwo & & \textbf{15.49} (0.37) & \textbf{41.39} (1.73) & \textbf{51.65} (2.30) & \textbf{62.01} (2.11)\\
        & \settingthree & & 15.23 (0.18) & 38.48 (4.11) & 49.03 (5.57) & 58.98 (4.74)\\
        \midrule
        \multirow{3}{*}{\settingthree} 
        & \settingone & \multirow{3}{*}{500} & 10.18 (0.20) & 0.09(0.10) & 0.19(0.20) & 10.33(0.30)\\
        & \settingtwo & & 11.82 (0.24) & 20.86 (1.43) & 27.28 (1.96) & 35.89 (1.81)\\
        & \settingthree & & \textbf{12.66} (0.12) & \textbf{37.54} (4.09) & \textbf{47.96} (5.43) & \textbf{55.40}(4.65)\\
        \bottomrule
    \end{tabular}}
    \vspace{-0.2cm}
    \caption{End-to-end cross evaluation of the full task}
    \vspace{-0.1cm}
    \label{tab:cross-full}
\end{table}
\newpage

\begin{table*}
  \centering
  \resizebox{\textwidth}{!}{
  \begin{tabular}{c|c|cccc|ccc}
    \toprule
    \multirow{2}{*}{\makecell{\# Training\\dialogs}} & \multirow{2}{*}{Model} & \multicolumn{4}{c|}{TOD Evaluation} & \multicolumn{3}{c}{ODD Evaluation}\\
    & & BLEU & Success & Inform & Combined & Accuracy & Success Rate & BLEU\\
    \midrule
    \multirow{3}{*}{200} & \taskbot & $16.36 (0.32)^{**}$ & $36.93 (5.46)^*$ & $48.19 (7.15)$ & $58.92 (6.19)$ & $0.00 (0.00)$ & $0.00 (0.00)$ & $1.25 (0.24)^{**}$\\
    & \chatbot & $0.91 (0.12)^{**}$ & $0.60 (0.00)$ & $10.71 (0.00)$ & $6.56 (0.12)^{**}$ & $\textbf{99.97} (0.05)$ & $\textbf{99.90} (0.15)$ & $7.57 (0.41)^{**}$\\
    & \mixedbot & $\textbf{16.37} (0.25)^{**}$ & $\textbf{41.29} (1.69)^{**}$ & $\textbf{53.61} (2.59)^{**}$ & $\textbf{63.85} (2.16)^{**}$ & $99.21 (0.50)^*$ & $98.00 (1.21)^*$ & $\textbf{7.75} (0.19)^{**}$\\
    \midrule
    \multirow{3}{*}{500} & \taskbot & $\textbf{17.73} (0.34)^{**}$ & $39.95 (3.22)$ & $50.28 (4.04)$ & $62.85 (3.54)$ & $0.00 (0.00)$ & $0.00 (0.00)$ & $1.09 (0.16)^{**}$\\
    & \chatbot & $0.83 (0.12)^{**}$ & $0.60 (0.00)$ & $10.70 (0.00)$ & $6.48 (0.12)^{**}$ & $\textbf{100.00} (0.00)$ & $\textbf{100.00} (0.00)$ & $\textbf{9.29} (0.18)^{**}$\\
    & \mixedbot & $17.50 (0.22)^{**}$ & $\textbf{42.69} (2.32)^*$ & $\textbf{54.11} (3.23)^*$ & $\textbf{65.90} (2.59)^*$ & $99.79 (0.16)$ & $99.42 (0.41)$ & $9.25 (0.20)^{**}$\\
    \bottomrule
  \end{tabular}}
  \caption{End-to-end evaluation of single tasks in the \settingone{} setting using GODEL as backbone. Almost all differences between GODEL-based models and T5-based models are statistically significant. (*p<0.05, **p<0.01).}
  \vspace{-0.5cm}
  \label{tab:TOD+chat_1}
\end{table*}
\begin{table*}
  \centering
  \resizebox{\textwidth}{!}{
  \begin{tabular}{c|c|cccc|ccc}
    \toprule
    \multirow{2}{*}{\makecell{\# Training\\dialogs}} & \multirow{2}{*}{Model} & \multicolumn{4}{c|}{TOD Evaluation} & \multicolumn{3}{c}{ODD Evaluation}\\
    & & BLEU & Success & Inform & Combined & Accuracy & Success Rate & BLEU\\
    \midrule
    \multirow{3}{*}{200} & \taskbot & $\textbf{16.48} (0.22)^{**}$ & $\textbf{38.79} (5.58)^{**}$ & $\textbf{50.78} (7.64)$ & $\textbf{61.26} (6.50)^*$ & $0.00 (0.00)$ & $0.00 (0.00)$ & $1.17 (0.12)^{**}$\\
    & \chatbot & $1.19 (0.13)^{**}$ & $0.60 (0.00)$ & $10.70 (0.00)$ & $6.84 (0.13)^{**}$ & $\textbf{100.00} (0.00)$ & $\textbf{100.00} (0.00)$ & $\textbf{6.04} (0.25)^{**}$\\
    & \mixedbot & $16.47 (0.37)^{**}$ & $34.93 (6.31)$ & $45.56 (8.24)$ & $56.71 (7.34)$ & $97.38 (0.50)^{**}$ & $93.22 (1.26)^{**}$ & $5.71 (0.20)^{**}$\\
    \midrule
    \multirow{3}{*}{500} & \taskbot & $\textbf{17.72} (0.33)^{**}$ & $42.46 (2.44)^{**}$ & $\textbf{53.53} (2.99)^{**}$ & $65.71 (2.74)^{**}$ & $0.00 (0.00)$ & $0.00 (0.00)$ & $1.00 (0.11)^{**}$\\
    & \chatbot & $1.11 (0.07)$ & $0.60 (0.00)$ & $10.70 (0.00)$ & $6.76 (0.07)^{**}$ & $\textbf{100.00} (0.00)$ & $\textbf{100.00} (0.00)$ & $\textbf{6.79} (0.25)^{**}$\\
    & \mixedbot & $17.71 (0.43)^{**}$ & $\textbf{42.69} (1.82)^{**}$ & $53.40 (2.35)$ & $\textbf{65.75} (2.16)^*$ & $98.65 (0.12)^{**}$ & $96.67 (0.31)^{**}$ & $6.75 (0.14)^{**}$\\
    \bottomrule
  \end{tabular}}
  \caption{End-to-end evaluation of single tasks in the \settingtwo{} setting using GODEL as backbone. Almost all differences between GODEL-based models and T5-based models are statistically significant. (*p<0.05, **p<0.01).}
  \label{tab:TOD+chat_2}
\end{table*}
\begin{table*}
  \centering
  \resizebox{\textwidth}{!}{
  \begin{tabular}{c|c|cccc|ccc}
    \toprule
    \multirow{2}{*}{\makecell{\# Training\\dialogs}} & \multirow{2}{*}{Model} & \multicolumn{4}{c|}{TOD Evaluation} & \multicolumn{3}{c}{ODD Evaluation}\\
    & & BLEU & Success & Inform & Combined & Accuracy & Success Rate & BLEU\\
    \midrule
    \multirow{3}{*}{200} & \taskbot & $\textbf{16.18} (0.31)^{**}$ & $\textbf{38.69} (6.25)$ & $\textbf{50.63} (7.32)$ & $\textbf{60.84} (6.69)$ & $0.00 (0.00)$ & $0.00 (0.00)$ & $0.91 (0.08)^{**}$\\
    & \chatbot & $1.14 (0.04)^{**}$ & $0.60 (0.00)$ & $10.70 (0.00)$ & $6.79 (0.04)^{**}$ & $\textbf{100.00} (0.00)$ & $\textbf{100.00} (0.00)$ & $\textbf{6.15} (0.40)^{**}$\\
    & \mixedbot & $16.04 (0.18)^{**}$ & $34.40 (5.55)$ & $45.04 (5.63)$ & $55.76 (5.52)$ & $98.22 (0.41)^{**}$ & $85.37 (3.07)^{**}$ & $5.91 (0.37)^{**}$\\
    \midrule
    \multirow{3}{*}{500} & \taskbot & $\textbf{17.40} (0.23)$ & $39.19 (3.33)$ & $49.83 (3.67)$ & $61.90 (3.57)$ & $0.00 (0.00)$ & $0.00 (0.00)$ & $0.90 (0.07)^{**}$\\
    & \chatbot & $1.04 (0.07)^{**}$ & $0.60 (0.00)$ & $10.70 (0.00)$ & $6.69 (0.07)^{**}$ & $\textbf{100.00} (0.00)$ & $\textbf{100.00} (0.00)$ & $\textbf{7.17} (0.11)^{**}$\\
    & \mixedbot & $17.26 (0.24)^{*}$ & $\textbf{40.69} (3.66)^{**}$ & $\textbf{51.94} (4.99)^*$ & $\textbf{63.57} (4.12)^{**}$ & $99.05 (0.38)$ & $91.86 (2.98)$ & $7.12 (0.12)^{**}$\\
    \bottomrule
  \end{tabular}}
  \caption{End-to-end evaluation of single tasks in the \settingthree{} setting using GODEL as backbone. Almost all differences between GODEL-based models and T5-based models are statistically significant. (*p<0.05, **p<0.01).}
  \vspace{-0.5cm}
  \label{tab:TOD+chat_3}
\end{table*}
\begin{table*}
    \centering
    \resizebox{\textwidth}{!}{
    \begin{tabular}{c|c|c|cccc|ccc}
        \toprule
        \multirow{2}{*}{\makecell{Evaluation\\setting}} & \multirow{2}{*}{\makecell{Training\\setting}} & \multirow{2}{*}{\makecell{\# Training\\dialogs}} & \multicolumn{4}{c|}{TOD Evaluation} & \multicolumn{3}{c}{ODD Evaluation}\\
        & & & BLEU & Success & Inform & Combined & Accuracy & Success Rate & BLEU\\
        \midrule
        \multirow{3}{*}{\makecell{init\\ODD}} 
        & \settingone & \multirow{3}{*}{500} & 17.50 (0.22) & \textbf{42.69}(2.32) & \textbf{54.11}(3.23) & \textbf{65.90}(2.59) & \textbf{99.79}(0.16) & \textbf{99.42}(0.41) & \textbf{9.25}(0.20)\\
        & \settingtwo & & \textbf{17.84}(0.43) & 40.49(2.38) & 51.30(3.06) & 63.74(2.66) & 91.65(8.24) & 77.54(20.89) & 4.66(0.24)\\
        & \settingthree & & 17.44(0.26) & 36.63(4.04) & 46.73(5.45) & 59.11(4.59) & 99.30(1.20) & 97.93(3.52) & 5.41(0.27)\\
        \midrule
        \multirow{3}{*}{\makecell{domain\\transition}}
        & \settingone & \multirow{3}{*}{500} & 17.08(0.37) & \textbf{43.41}(2.73) & \textbf{55.03}(4.11) & \textbf{66.30}(3.29) & 35.67(14.32) & 4.26(3.34) & 2.33(0.33)\\
        & \settingtwo & & \textbf{17.71}(0.43) & 42.69(1.82) & 53.40(2.35) & 65.75(2.16) & 98.65(0.12) & 96.57(0.31) & 6.75(0.14)\\
        & \settingthree & & 17.28(0.19) & 38.83(4.11) & 49.55(5.57) & 61.47(4.76) & \textbf{99.58}(0.17) & \textbf{98.91}(0.43) & \textbf{7.22}(0.21)\\
        \midrule
        \multirow{3}{*}{\makecell{multiple\\ODDs}} 
        & \settingone & \multirow{3}{*}{500} & 16.44(0.30) & 39.46(2.91) & 51.50(3.62) & 61.92(3.10) & 31.28(14.11) & 0.57(0.44)	& 2.21(0.30)\\
        & \settingtwo & & 17.15(0.43) & 38.80(1.13) & 50.06(1.46) & 61.58(1.39) & 93.04(0.79) & 53.91(4.02) & 5.39(0.09)\\
        & \settingthree & & \textbf{17.26}(0.24) & \textbf{40.69}(3.66) & \textbf{51.94}(4.99) & \textbf{63.57}(4.12) & \textbf{99.05}(0.38) & \textbf{91.86}(2.98) & \textbf{7.12}(0.12)\\
        \bottomrule
    \end{tabular}}
     \caption{End-to-end cross evaluation of single tasks}
     \label{tab:cross-tod-odd}
\end{table*}

\end{document}